\documentclass[conference]{IEEEtran}
\usepackage{cite}
\usepackage{graphicx}
\usepackage{amsmath}
\usepackage{algorithmic}
\usepackage[utf8]{inputenc}

\begin{document}

\title{Controlling the Charging of Electric \\ Vehicles with Neural Networks}

\author{\IEEEauthorblockN{Martin Pilát}
\IEEEauthorblockA{Charles University, Faculty of Mathematics and Physics\\
Malostranské náměstí 25, 118 00 Prague, Czech Republic\\
Email: Martin.Pilat@mff.cuni.cz}
}

%TODO:
% R1
% 1. IGNORE: Hence it does not explain why a household would buy and use these controllers? 
% 2. DONE: I also did not understand how the type H controllers can reduce the global objective function and How they are different than random charging at low consumption times in having estimate of electricity profile.
% R2
% 3. The author could put some pictures to illustrate his proposal.
% 4. DONE: The use of the low-pass filter in the output module helps to remove  possible fluctuations in the output of the control -> is necessary?
% 5. DONE: We could also use the optimization to train the controllers to minimize the losses in the grid..." - Only contribution:  the network losses are not always the best for the system, it may be that at other points in the  network the voltage or charge profile is above the permitted limits.

\maketitle

\begin{abstract}
We propose and evaluate controllers for the coordination of the charging of electric vehicles. The controllers are based on neural networks and are completely de-centralized, in the sense that the charging current is completely decided by the controller itself. One of the versions of the controllers does not require any outside communication at all. 

We test controllers based on two different architectures of neural networks -- the feed-forward networks and the echo state networks. The networks are optimized by either an evolutionary algorithm (CMA-ES) or by a gradient-based method. The results of the different architectures and the different optimization algorithms are compared in a realistic scenario. We show that the controllers are able to charge the cars while keeping the peak consumptions almost the same as when no charging is performed. Moreover, the controllers fill the valleys of the consumption thus reducing the difference between the maximum and minimum consumption in the grid.
\end{abstract}

% no keywords

\section{Introduction}

The sales of electric vehicles are quickly growing around the world, and while in most countries the market share of electric vehicles is still only around 1 percent, in Norway, due to government subsidies, it is more than 30 percent according to the data provided by the European Alternative Fuels Observatory\footnote{http://www.eafo.eu/eu}. At the same time, some countries have already indicated that they want to replace the cars powered by fossil fuel by cars using an alternative clean source of energy in the following decades.

According to Eurostat data~\cite{eurostat}, there are currently over 200 million of cars registered in Europe. If all of these were replaced by electric vehicles, the power needed for their charging would be almost equal to the consumption of all households. While this amounts to only a quarter of the overall electricity consumption in the EU, we can assume that that consumption would be centered around residential areas, and in these it could in fact double the consumption. The problem is even worse, as most people would come home from work at similar time in the afternoon, and if they all start charging their cars at the same time, the charging can cause significant peaks in the power consumption. 

This shows the importance of coordination in the charging of electric vehicles and the problem has been tackled in the literature from different points of view and several different methods to solve it have been proposed. For example, Deilami \emph{et al.}~\cite{5986769} provide an algorithm for real-time coordination of electric vehicle charging and aim to minimize the losses that the charging can bring. Gan \emph{et al.}~\cite{gan2013optimal} provide a charging protocol and prove that it is optimal under some assumptions (like all the charging requests are known beforehand). In the protocol, the grid provides a control signal to the controllers and they solve an optimization problem in order to compute a new charging profile. In the algorithm by Ma \emph{et al.}~\cite{6081962}, the vehicles exchange the charging requests and their predictions of consumption in an interactive way  with the utility company in order to find an optimal charging strategy for all vehicles. There are also algorithms which aim to solve the problem centrally~\cite{4839973} on a single node that controls the charging of all the vehicles.

While most of the above mentioned algorithms are de-centralized in the sense that (parts of) the optimization problems are solved by the chargers themselves, they still require quite frequent communication with a coordinating third party. This means that the communication infrastructure must be built and it also reduces the privacy of the owners of the electric vehicles -- not only is the presence of the electric vehicle shared with a third party but also the whole charging request including the time when the vehicle must be charged is shared. Many people are concerned with their privacy and such sharing may stop them from using the smart controllers.

Therefore, in this paper, we aim to provide controllers, that are completely de-centralized and require no or minimal communication with third parties. Moreover, we seek controllers that are both memory and computationally efficient and thus can be implemented in cheap embedded systems. We believe neural networks can provide a base for such controllers as only their weights need to be stored and once they are trained, they are extremely fast to evaluate. 

Thus, the main goals of this paper are to provide a general architecture for a neural network based controller and to evaluate this architecture with different types of neural networks. An important question is, how to train such controllers and we therefore compare two different algorithms for the training. The evaluation of the controllers is performed in a realistic scenario based on electricity consumption data from real households that own an electric vehicle and on travel data from the National Transportation Travel Survey. 

We have already presented some ideas and preliminary results on electric vehicle charging with neural networks in a short paper~\cite{evostar}. Compared to that paper, here we add new features for the controllers, and more importantly add new controllers based on echo state networks. We also provide a gradient-based optimization algorithm for the training of the parameters of the controllers in addition to the evolutionary algorithm used in the above-mentioned paper.

\section{Creating Neural Network Controllers}

We assume that each household is equipped with a smart charger that provides the charging current for the vehicle in each point in time. Specifically, we assume that once the owners of the electric vehicles come home they specify the time the car needs to be charged, thus specifying a charging request that, apart from the time when the car needs to be charged, also contains the required charge (can be computed from the state of the battery) and the maximum charging current (given by the technical specification of the car). The controller then makes sure the car will be charged in time while optimizing the electricity consumption of the whole grid.

We consider two types of controllers defined by the data they receive from the grid. Controllers of type H do not receive any information from the grid and therefore use only information that is available locally in the household. On the other hand, the controllers of type A receive information about the overall consumption in the grid and additionally can also use the information from the household. Intuitively, the type A controller should provide better global results, as they have access to the information about the overall consumption in the grid. The type H controllers can only react to changes in electricity consumption in each household, which (while typically correlated with the overall grid consumption) does not provide such a direct connection with the optimization objective. On the other hand, the type H controllers are simpler and do not require any outside communication.

For the ease of description, we divide the controller into three modules -- the input module, control module and output module. The input module receives the raw inputs from the environment and transforms them into the inputs of the control module, the control module provides a raw control output based on these inputs and, finally, the output module transforms the raw control output to the charging speed that will be used in the next time step.

We assume the controllers in each household are identical, but they also have an internal state that can be different depending on the consumption of each household, its charging request and so on. The controller for a household $h$ can thus be expressed as a function $C(x_h, x_g, \theta, \Sigma_h)$, where $x_h$ are the inputs available for the household $x_g$ are the inputs available from the grid ($x_g = \emptyset$ for controllers of type H), $\theta = (\theta_i, \theta_c, \theta_o)$ are the parameters of the input, control and output modules, and $\Sigma_h^t = (\Sigma_{h,i}^t,\Sigma_{h,c}^t, \Sigma_{h,o}^t)$ is the internal state of the controller for household $h$ at time $t$ consisting of the internal states of the modules of the controller. The controller returns the charging speed and new internal state.

\subsection{Input module}

In each step, the input module reads the information available in the household -- the current time, the current charging request, the overall electricity consumption of the household and the charging consumption of the household. Controllers of type A also receive the information on the overall consumption of the grid from a central node. Based on this information, the input module computes the inputs for the control module. These are 
\begin{enumerate}
	\item the current time of day encoded as two numbers -- $\cos(2\pi \frac{t}{86,400})$ and $\sin(2\pi \frac{t}{86,400})$, where $t$ is the time of day measured in seconds since midnight,
	\item weekend flag -- 0 if the current day is a Saturday or Sunday and 1 otherwise,
	\item the percentage of steps remaining from the charging request, 
	\item the percentage of the charging request that is not yet fulfilled,
	\item the minimum charging speed required to charge the car on time divided by the maximum charging speed,
	\item the constant charging speed that would charge the car on time if used for the rest of the request divided by the maximum charging speed,
	\item the current consumption of the household divided by the average consumption of the household over the last 24 hours,
	\item the change in the household consumption compared to the last time step, to one hour ago, and to three hours ago (3 inputs),
	\item a number expressing how the current household consumption compares to the maximum and minimum consumption in the last 24 hours, computed as $\frac{\mathrm{current}-\mathrm{min}}{\mathrm{max}-\mathrm{min}}$, and 
	\item for controllers of type A, the same information as in points 7-9 but with the overall grid consumption instead of the household consumption.
\end{enumerate}

The encoding of time using the $\sin$ and $\cos$ ensures that for similar times the inputs will be similar, even for times around midnight when the input $t$ is non-continuous. We distinguish only between workday and weekend as the whole training is currently performed only on a few weeks and therefore making difference between various workdays seems unnecessary.

Overall, controllers of type H compute 12 features, while controllers of type A compute 17 features. The internal state of the input module consists of the last consumptions for the last 24 hours, which are needed to compute some of the features. There are currently no tunable parameters in the input module and thus $\theta_i = \emptyset$. 

\subsection{Control module}

The control module receives the pre-processed information from the input module and outputs a raw charging request. This output is a number between 0 and 1 and expresses the charging speed as the percentage of the maximum possible charging speed.

The control module first updates its internal states depending on its inputs. Then, if no charging request is active at the time, it directly returns 0, otherwise it computes the raw charging speed.

We consider two types of control modules in this work. One of them is a simple feed-forward neural network (called simply NN in the rest of the paper) and the other is an echo state network (ESN). The NN control module is purely reactive and does not have any internal state. It computes its raw output $r$ based on the inputs $\vec{x}$ it receives from the input module as $$r = \mathrm{sigm}(W_2\mathrm{ReLU}(W_1\vec{x} + \vec{b}_1) + \vec{b}_2)\,,$$ where $\mathrm{sigm}(x) = \frac{1}{1+e^{-x}}$ is the logistic sigmoid, $\mathrm{ReLU}(x) = \max(0,x)$ is the linear rectifier function, and $W_1, W_2, \vec{b}_1, \vec{b}_2$ are the weight matrices and biases in the network. In this case, the parameters of the control module are $\theta_c = (W_1, W_2, \vec{b}_1, \vec{b}_2)$. The size of the hidden layer is a meta-parameter of the NN control module and is set from the outside.

The ESN controllers are based on echo state networks~\cite{jaeger2001echo}, which are an example of the idea of reservoir computing. As such they should provide better performance for problems where the inputs are correlated and where the outputs affect the next inputs as is the case of the problem at hand. They are basically recurrent neural networks, where the recurrent part is generated randomly and is not optimized in any way. The recurrent part -- also called reservoir -- is commonly represented by a single random matrix. In the ESN control module, the internal state $\sigma$ is represented by the state of the echo state network. This state must be first updated according to the equation $$\vec{\sigma}_t \gets (1-\alpha) \vec{\sigma}_{t-1} + \alpha \tanh(W_r \vec{\sigma}_{t-1} + W_{in} \vec{x} + b_{in})\,,$$ where $\sigma_{t-1}$ and $\sigma_t$ are the internal states at times $t-1$ and $t$, $\vec{x}$ is the input vector from the input module, $W_{in}$ and $b_{in}$ are the input weight matrix and input bias, $W_r$ is the reservoir matrix, and $\alpha$ is a parameter that affects how the new state is affected by the old one. This update in the control module must be performed always, regardless of whether a charging request is active or not. If a charging request is not active, the inputs that rely on the charging requests are set to 0. In this case, the output of the control module is also set to 0. If a request is active, the output is computed as $\mathrm{sigm}(W_{out} (\vec{\sigma}, \vec{x}) + b_{out})$ where $(\vec{\sigma_t}, \vec{x})$ is the concatenation of the internal state $\vec{\sigma}$ and the input vector $\vec{x}$, $W_{out}$ is the output weight matrix and $b_{out}$ is the output bias.

Although the ESN control module contains a number of parameters, only the output weight $W_{out}$ and the output bias $b_{out}$ are optimized, the rest is fixed. Therefore the parameters $\theta_{c} = (W_{out}, b_{out})$. The parameter $\alpha$ as well as the size of the reservoir matrix are hyper-parameters set from the outside, the rest of the matrices and biases are generated randomly and remain fixed during the optimization.

\subsection{Output module}

The output module obtains the raw charging output from the control module and first multiplies it by the maximum charging speed to obtain a raw charging current. Then it works as a low-pass filter, i.e. it computes a weighted sum of the current raw charging current with the previous output. Finally, it ensures the charging request will be fulfilled, i.e. it makes sure that the charging speed is at least he minimum charging speed required to charge the vehicle in time. Overall, the output module can be expressed as $o_t = \min(s_{min}, \beta o_{t-1} + (1-\beta)s_{max} r_t)$, where $s_{min}$ is the minimum required charging speed, $s_{max}$ is the maximum charging speed, $r_t$ is the raw output of the control module at time $t$ and $o_t$ is the output of the output module at time $t$. The internal state of the module is thus only its last output, and it also has a single internal parameter $\beta$.

The use of the low-pass filter in the output module helps to remove possible fluctuations in the output of the control module and in general makes the changes of the outputs slower. During preliminary experiments, we found out that it slightly improves the performance of the controllers and, more importantly, limits the risk of large fluctuations caused by the controllers.

\subsection{Optimization}

It is simple to implement a simulation of the charging of electric vehicles for given charging requests $R$, baseline electricity consumption $B$ (without the electricity needed for charging) and given internal parameters $\theta$ of the controller $C(x_h, x_g, \theta, \Sigma_h)$. The simulator first initializes the internal states of the controllers of each household $h$ ($\Sigma^t_h$) to zeros. Then in each time step $t$ it first updates all the active charging requests -- it adds new request for vehicles that became available for charging, decreases the remaining required charge for the existing requests (according to the charging speeds from the previous step), and removes finished requests. Finally, it calls the controller $C(x_h, x_g, \theta, \Sigma^{t-1}_h)$ for each household $h$ to obtain the new charging speed $c^h_t$ and new internal state $\Sigma^t_h$. 

The output of the simulation is for each time step $t$ the consumption of the whole grid $c_t = B_t + \sum_h c_t^h$, where $B_t$ is the uncontrollable baseline load in the grid and $c_t^h$ are the charging speeds for each household $h$. The function implementing the simulation and returning the vector $\vec{c}$ of consumptions $c_t$ will be called $\mathrm{sim}(R, B | \theta)$.

We can now in principle use any optimization algorithm in order to set the parameters $\theta$ in such a way that an arbitrary objective $o(c)$ computed from the vector of consumptions is optimized. For example, a common objective is the minimization of the peaks in the consumption, which can be expressed as $\min_\theta \max_t c_t$. We could also use the optimization to train the controllers to optimize other criteria regarding the loads in the grid, such as minimizing the electricity price or minimizing the losses in the grid. As the losses depend on the square of the consumptions, the objective in this case would be $\min_\theta \sum_t c_t^2$. All of these optimization problems have quite a large number of constraints that ensure the charging speed is always less than the maximum charging speed for the given household and also ensure that the sum of the charges matches the required charge for each charging request. 

In this work, we decided to minimize the standard deviation of the consumptions, which is in fact equivalent to the minimization of losses because the sum of consumptions $\sum_t c_t$ is fixed by the charging requests and the baseline load $B$. Therefore, the optimization problem we aim to solve is $\min_\theta \mathrm{std}(\mathrm{sim}(R, B | \theta))$. In order to solve the optimization problem, we compare two different algorithms -- the CMA-ES evolution strategy~\cite{hansen2001completely} and a gradient method. 

The CMA-ES is currently considered one of the best algorithms for continuous optimization and its main advantage for the problem at hand is that it does not need the gradients of the objective. 

On the other hand, neural networks are typically trained using gradient algorithms. In this case, the gradient $\nabla_\theta \mathrm{std}(\mathrm{sim}(R, B | \theta))$ needs to be computed numerically, which leads to a single run of simulation for each of the parameters $\theta$. As there are usually at least tens of parameters, this is extremely time consuming. In order to speed the computation of the gradient up, we use shorter simulations starting at random points in time during the computation of the gradient, thus effectively implementing a variant of stochastic gradient descend. After the gradient is computed the parameters $\theta$ are updated as usual $$\theta \gets \theta - \lambda \nabla_\theta \mathrm{std}(\mathrm{sim}(R, B | \theta))\,,$$ where $\lambda$ denotes the step size. In order to set the step size, we could use line search in the direction of the gradient, as in BFGS, but such gradient-based line search is extremely slow and therefore we instead try a few different values for $\lambda$ and choose the one that leads to the best value of the objective.

As the output module has only one parameter, we actually optimize it separately after the rest of the parameters were set using one of the methods mentioned above. This is motivated by the fact the parameter $\beta$ of the output module actually limits the effect of the other parameters on the result of the simulation (in an extreme case of $\beta = 1$ the output of the control module is completely ignored).

\section{Experiments}

In this section, we describe the experiments we performed and their results. We first show, how we obtained realistic data for the simulation, then we provide the parameters of the experiments and finally we discuss their results. 

\subsection{Data preparation}

In order to test the controllers, we need realistic data on electricity consumption and realistic charging requests. In order to create such data, we combined the information from the Pecan Street's Dataport (PSD) database~\cite{dataport} and from the National Household Transportation Survey (NHTS)~\cite{nhts}.

The PSD database contains detailed information on the electricity consumption of several hundred households, mostly from Texas. The available information consists of the electricity consumption of each electrical appliance in the household. Among these are electric cars and the current needed for their charging. Moreover, the overall consumption of each household is also available. While the information is available with 1-minute resolution, we used 15-minute resolution for the experiments in order to speed them up. Specifically, we used data from the first three months of 2015 for the households in Texas that have an electric vehicle (there are 74 such households). 

Unfortunately, the data from the PSD database cannot be used to detect when a car was connected to the grid and thus available for charging. Therefore, we used the NHTS data that contain information on travels performed by US households. From these data, for each household, we selected the first travel by car in the day starting at home and the last travel by car in the day ending at home. For each of these travels we also recorded the day of week. In this way, we obtained realistic times, when cars are not at home and therefore not available for charging. We again used data from Texas from the first three months of the year 2009 (newer data are currently not available). In this way, we obtained over 3,000 different travel records. 

In order to obtain realistic charging requests for the simulation, the data from the PSD database and NHTS survey were combined. First, for each household from the PSD database and each day we take a random travel from the records from the NHTS with a matching day of the week. This gives for each day the time when the car is not present. From these data, the charging requests are created. Each charging request starts at the time the car returns home and ends at the time the car leaves on the next day. The maximum charging current is the maximum charging current for the given household as observed in the PSD data. The most complex part is computing the required charge -- it is the charge used in the PSD data between the time the car leaves on days $d$ and $d+1$. This way we ensure that the total charging in the simulation is the same regardless of the times of arrival and departure in the requests. If we only took the charge that is used in the PSD during the time of the active request, some of the charging would be missed, as the two databases are not aligned in time (i.e. there can be active charging in the PSD data at the same time the car is not available according to the generated travel for the given day). After these steps, some of the charging request will have an empty charging requirement, such requests are ignored in the simulation.

The above mentioned steps ensure that the charging requests are realistic -- the times of arrival and departure have the same distribution as the travels recorded in the NHTS survey for households in a similar geographic area and on the same days of the week. The requested amount of charge is the same as the amount that was actually used in the given household for charging. While we generate a travel for each day, the fact that empty charging requests are ignored also means that travel on days where no charging in the PSD database was performed are ignored in the simulation. Thus, in the simulation, the cars charge only on days when charging was required in the PSD database.

\subsection{Experiment settings}

For the optimization of the controllers we used data for 15 days in the beginning of January 2015 (starting on January $2^{nd}$ at 4 p.m.). The consumptions for the first 24 hours of each simulation are not included in the objective and are used only to initialize the internal state of the controllers. 

For the NN-based controllers, the initial weights in the network are set to $0.1 \cdot\mathcal{N}(0,1)$. The hidden layer contains 5 neurons. For the ESN-based controllers, the input weight in the ESN are generated randomly between $-0.5$ and $0.5$. The recurrent weights are first also generated from the same distribution, but then 90 percent of random values are removed from the matrix in order to increase its sparsity and the whole matrix is divided by its spectral radius. The weight of the previous state in the update of the internal state is set to 0.9, the new state has weight of 0.1. The reservoir size is set to 100 neurons.

For the optimization of the controller we chose two algorithms, either the CMA-ES or a gradient based one. The CMA-ES\cite{hansen2001completely} algorithm is run for 250 generations with the population of 16 individuals. The selected population size is close to the recommended value (which depends on the number of parameters in the optimization problem). Moreover, we used parallel evaluation of the fitness function with 8 CPUs, and for the CPU utilization it is better if the population size is a multiple of the number of CPUs.

As we mentioned, for the gradient method, the gradient of the objective is computed numerically. As each gradient evaluation is time consuming, we run the simulation for the gradient computation only for 72 hours starting randomly in the training interval and compute the objective based on the last 48 hours of this run. After the gradient is computed, we need to set the step size $\lambda$. We compute the objective for 8 different step sizes and use the one that minimizes the objective. These 8 values are $\lambda \in \{0.0001, 0.0005, 0.001, 0.005, 0.01, 0.05, 0.1, 0.5\}$. For the selection of the step size, the simulation is run for all the 15 training days. This gradient step is repeated 250 times.

During the optimization, the parameter $\beta$ of the low-pass filter is set to 0 and the filter is not used. After the optimization, the data from 8 days starting on January $19^{th}$ are used to optimize this parameter. We have previously observed that reasonable values of the parameter are rather low, so we try values of $\beta \in \{0.0, 0.1, 0.2, 0.3, 0.4, 0.6, 0.8, 1.0\}$ and use the best one according to the objective.

In order to compare the optimized controllers, we run another 3-week simulation starting on March $8^{th}, 2015$ and computed various criteria using the loads from this simulation. Again the loads during the first 24 hours are ignored in the simulation.

Apart from the NN and ESN controllers we also implemented three simple baseline controllers called ``max charge'', ``min charge'', and ``const charge''. The ``max charge'' controller always uses the maximum possible charging speed, thus it also simulates the situations where no smart controllers are used and the cars are charged as soon as possible. The ``min charge'' controller on the other hand moves the charging to the latest possible time as it uses the minimum charging speed required to charge the car at time. Finally, the ``const charge'' controller charges the cars with constant speed through the whole time the car is available for charging.

The names of the controllers as they are used in the rest of the paper consist of three parts and are of the form model-inputs+optimization. The ``model'' is either NN or ESN denoting the feed-forward neural network and echo state network respectively. The inputs are a single letter A or H, where A denotes  the models that use all the inputs (including those from the grid), and H denotes the models that use only the information available locally in the household. Finally, the optimization can be either CMA or Grad, with the former denoting that CMA-ES was used for the optimization and the latter denoting that the gradient method was used. Each of the experiments was repeated five times.

\subsection{Results}

\begin{figure}
\centering
\includegraphics[scale=0.9]{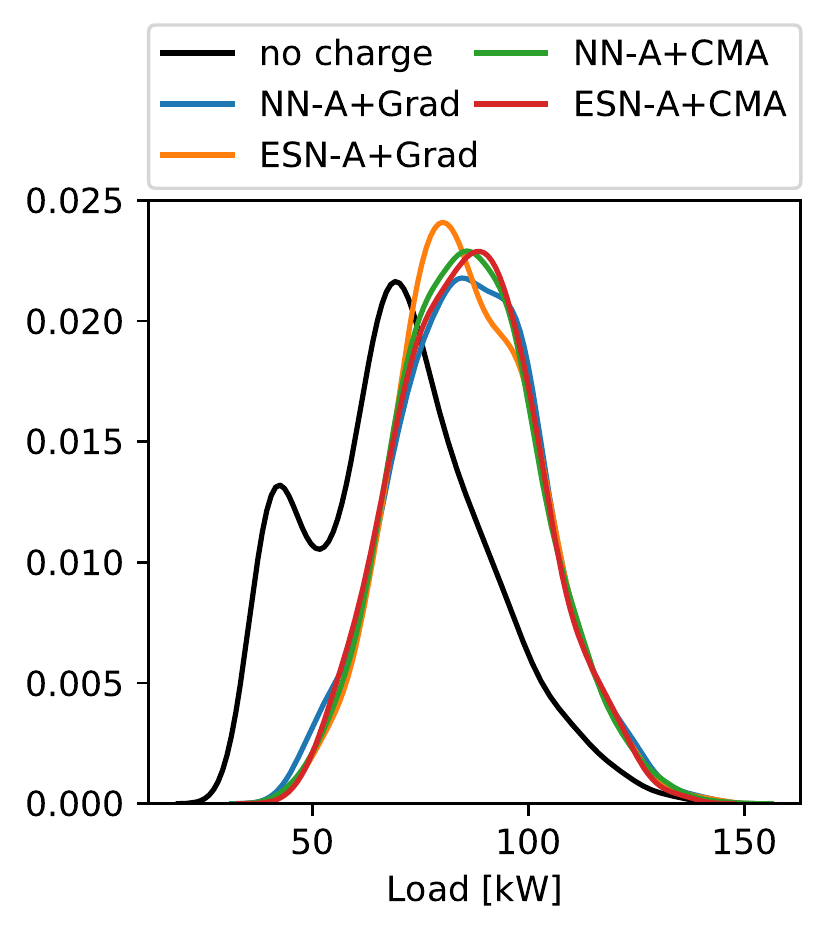}
\caption{The histogram of the loads in the network without charging and with four different controllers (the controllers with median objective from five independent runs are shown).}
\label{fig:hist}
\end{figure}

Before we get to the comparison of controllers, we will discuss how the charging affects the electricity consumption. In the data we used, charging of the electric vehicles account for approximately 18 percent of the overall consumption. The goal is to use the charging consumption to stabilize the overall consumption of the grid. Figure~\ref{fig:hist} shows the histogram of the consumptions without charging and with the controllers we found in the experiments (controllers with median objective for each of the types are displayed). Obviously, with the added charging the average consumption is higher and therefore the histogram is moved to the right. However, the maximum  consumption with and without charging is similar, i.e. the controllers are able to charge the cars without increasing the maximum consumption. At the same time, with the charging, the minimum consumption is increased and thus the whole grid operates in a narrower range of the loads, which simplifies the planning of power production. 

\begin{figure*}[t]
\centering
\includegraphics[scale=0.9]{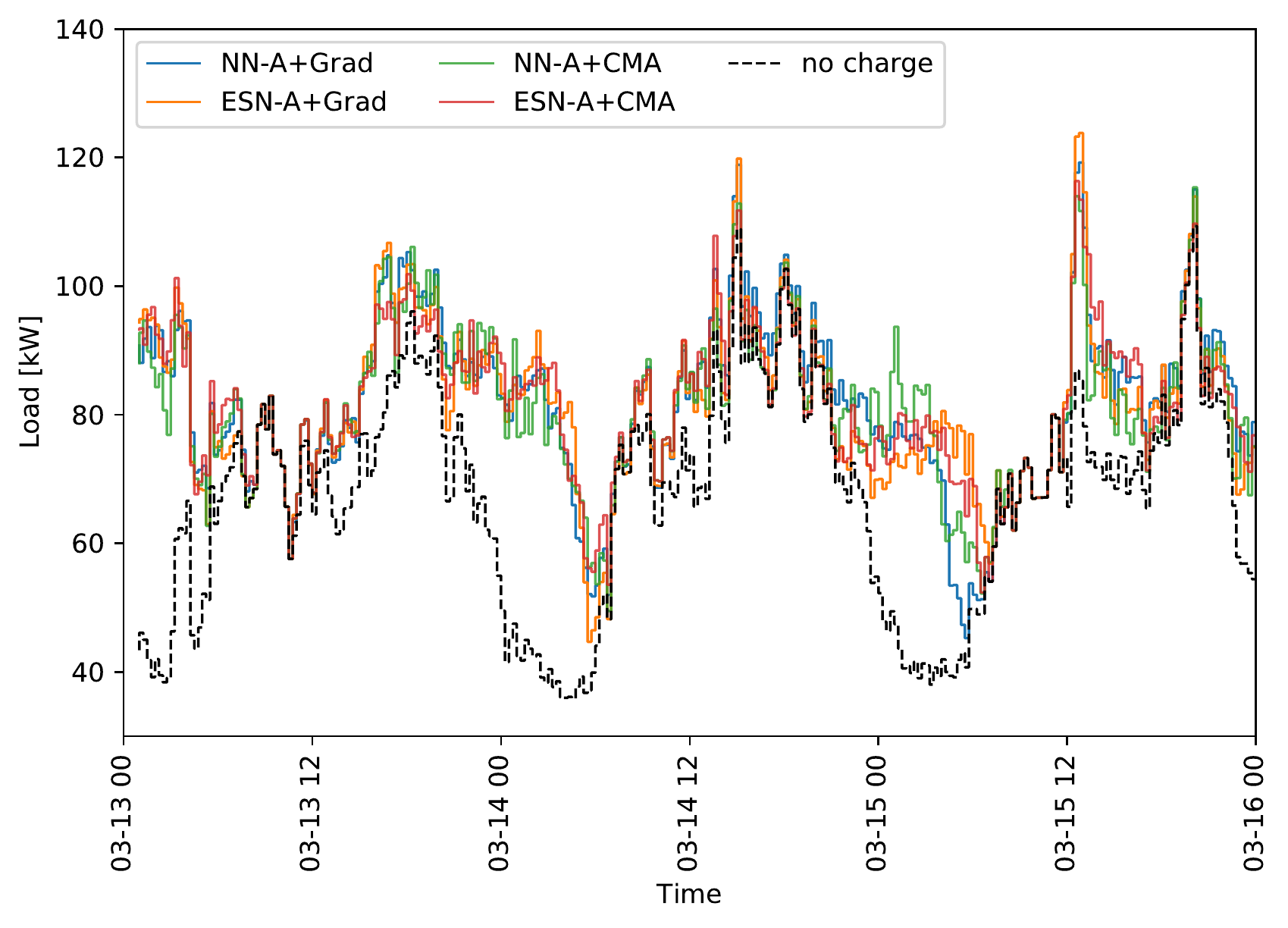}
\caption{Example of the electricity load for four different controllers during two full days in March 2015.}
\label{fig:consumption}
\end{figure*}

An example of how the load changes in time is presented in Figure~\ref{fig:consumption}. It shows the baseline load without charging and also the load obtained while some of the controllers were used for two full days in March 2015. In all cases, the presented controllers are those that had median value of the objective of the five runs. The differences between the different types of controllers are rather small. But we can see that the controllers optimized by the gradient method tend to make smaller changes between consecutive the time steps (we discuss this in more detail bellow). We can also see quite a large spike in the load around the noon on March 15. This was caused by a rather short and sudden drop followed by quick increase in the baseline load. These fast changes confused the controllers that reacted to the quick drop by increasing the charging speeds. Overall, during these two days, all the controllers were able to partially fill the valleys in the consumption, but they had some problems during the ends of the valleys in the morning.

\begin{table*}[t]
\centering
\caption{Comparison of the different types of controllers and optimization algorithms. The table shows the optimization objective (the standard deviation of the consumptions), the minimum load and maximum load over the three weeks in March 2015 and also the 2.5-th and 97.5-th percentile of the consumptions. The numbers are averages over five independent runs.}
\label{tab:results}
\begin{tabular}{lrrrrr}
\hline
Controller+Optimizer &  Objective &  Min. load &  2.5-th perc. &  97.5-th perc. &  Max. load \\
\hline
no charge 		& 20.130 & 32.919 & 37.632 & 112.486 & 134.981 \\
\hline
max charge 		& 34.720    & 32.919 & 38.035   & 161.692   & 194.111 \\
min charge		& 24.780    & 35.383 & 42.437   & 141.031   & 169.649 \\
const charge	& 19.451    & 49.812 & 58.442   & 133.924   & 155.238 \\
\hline
NN-A+CMA     	&     15.464 &    45.520 &  57.988 &  118.574 &   141.228 \\
NN-H+CMA     	&     16.065 &    49.341 &  58.969 &  120.957 &   142.251 \\
ESN-A+CMA    	&     15.190 &    48.076 &  59.187 &  118.842 &   139.792 \\
ESN-H+CMA    	&     16.920 &    48.454 &  58.479 &  123.046 &   150.728 \\
\hline
NN-A+CMA (long) &     15.197 &    47.495 &  59.151 &  118.346 &   137.393 \\
ESN-A+CMA (long)&     15.695 &    44.793 &  56.946 &  119.084 &   138.323 \\
\hline
NN-A+Grad 		&     16.851 &    43.644 &  55.044 &  121.834 &   143.242 \\
NN-H+Grad  		&     17.348 &    46.933 &  58.304 &  124.432 &   149.457 \\
ESN-A+Grad   	&     16.164 &    45.044 &  56.632 &  120.399 &   142.714 \\
ESN-H+Grad 	    &     17.641 &    44.297 &  58.635 &  126.203 &   150.466 \\
\hline
\end{tabular}
\end{table*}

The overview of the results of the experiments is presented in Table~\ref{tab:results}. The table shows the objective and basic statistics on the consumption for all the tested combinations of controllers and optimizers as well as for the baseline controllers. For comparison, we also added the same values for the consumption without any charging (as ``no charge'' in the table). The numbers are averages over five independent runs for each of the controllers.

We computed five different criteria for the controllers -- the optimization objective, i.e. the standard deviation of the consumptions of the whole grid which shall be minimized; the minimum and maximum load during the simulation (greater minimum and lower maximum are preferred) and also the 2.5-th and the 97.5-th percentile of the consumptions (again, the greater values of the former and lower values of the latter are better). The percentiles are added because the minimum and maximum can be reached under uncommon circumstances found in the data and may often be consequences of the baseline consumption rather than the charging. The percentiles remove some of these effects.

Interestingly, there are not very large differences among all the optimized controllers based on these criteria. According to the objective, the best results were achieved by the ESN-A+CMA controller, closely followed by the NN-A+CMA (and especially its version with longer run). The performance of these controllers is a clear improvement over the baseline controllers. For example, the ``max charge'' controller simulates the situation, when the charging starts immediately when the car is connected, the maximum load for this controller is 191 kW, while for the optimized controllers it is less than 140 kW. Similarly for the minimum loads -- the optimized controllers reach a minimum of 50 kW, while the baseline controller drops to 33 kW. These numbers should also be compared to the situation without any charging -- in such a case the maximum load is 135 kW and the 97.5-th percentile of the load is 112 kW, with the optimized controllers these numbers are only minimally increased to 140 kW and 119 kW respectively. This shows that the charging of electric vehicles can be easily performed in the grid without any need to expand its capacity if the controllers are used.

We can also make some more general observations. First of all, in most cases the performance of the controllers is better if the overall consumption of the grid is available as the input, the only exception here is the minimal load criterion, in which the H variants generally provide slightly better results. We believe this is caused by the fact that the H variants are less synchronized as they have no common inputs and therefore there is a higher chance that some of the households charge their cars in each time step. Moreover, the difference between the H and A variants is not very large and therefore the H variants may be practically interesting, because they can be incorporated into the grid without any need to invest into communication between the households and the grid.

As for the two optimization algorithms, it seems that for this particular problem the CMA-ES algorithm provides better results than the gradient-based optimization. The controllers optimized by the gradient method generally have higher maximum consumption with lower minimum consumption compared to the CMA-ES based controllers. For example, the NN-A+CMA controller operates most of the time between 58.0 kW and 118.5 kW, while the gradient version is between 55.0 kW and 121.8 kW. The range of the latter is thus approximately 6 kW (or 10 percent) wider.

For comparison, we also made longer runs for some of the controllers optimized by the CMA-ES algorithm (500 generations instead of 250). For the ESN-based controller the result of the longer run are actually slightly worse than the results for the shorter run. For the NN-A+CMA controller, the results are slightly better especially concerning the maximum consumption which was decreased from 141 kW to 137 kW. The performance of the NN-A+CMA controller after the longer run is actually close to the performance of the ESN-A+CMA controller after the shorter run.

\begin{figure}
\centering
\includegraphics[scale=0.9]{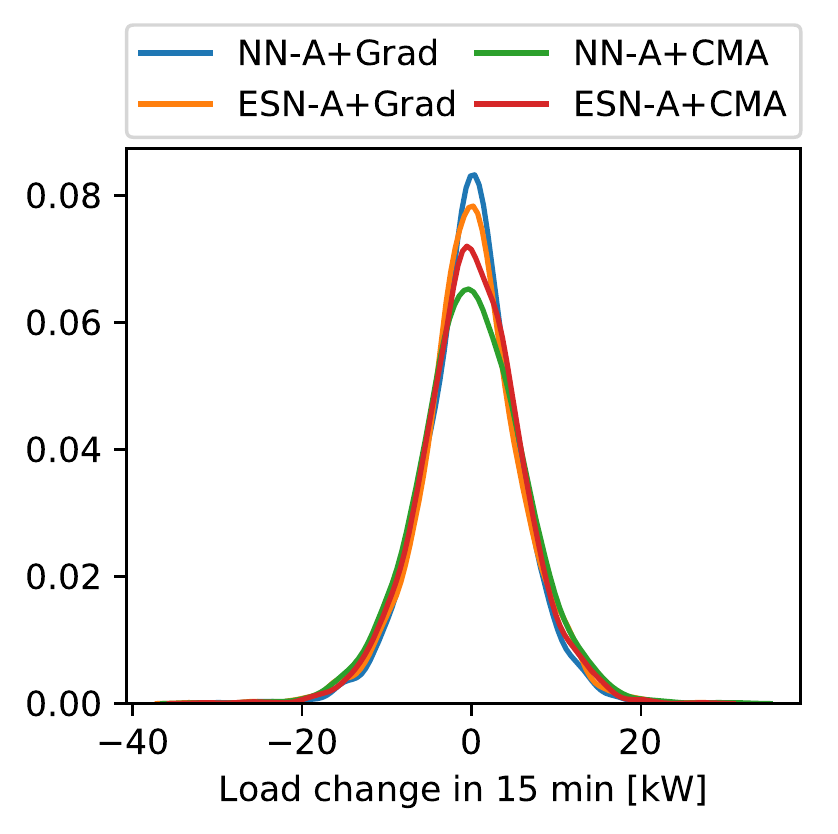}
\caption{The histogram of the changes of the loads in after each 15-minute step in the grid with the median controllers.}
\label{fig:changes_hist}
\end{figure}

We already noted that the gradient-based controllers tend to make smaller changes between consecutive time steps. This is graphically presented in Figure~\ref{fig:changes_hist} that shows the histogram of the changes for the four controllers (the same as in the plot of the load profile and of the histogram of consumptions). We can see that the peaks around 0 are higher for the gradient-based controllers, which indicates that the changes tend to be smaller. This would mean that while the gradient-optimized controllers have worse performance overall, they are more stable, in the sense, that the overall consumption changes more slowly. Which can again be an important practical consideration. Similar effect can also be observed by comparing the parameters of the low-pass filter. The gradient-based controllers seem to not need any low pass filter on the output, as in the tuning of this filter, the best value of the parameter $\beta$ was almost always 0. 

We observed a similar behavior for the ESN-based controllers -- they also work better for lower values of the parameter $\beta$ of the low-pass filter as well as making smaller changes according to the histogram. The effect was also observed during the optimization where in the initial generations, the ESN-based controllers had significantly better objective values than the NN-based ones. All of these effects may be caused by the reservoir which accounts for most of the inputs to the last layer of the controller. The reservoir changes only slowly and thus limits the changes of the controller output.

\section{Conclusions and Future Work}

We have designed two types of controllers for the electric vehicle charging problem. The controllers are de-centralized in the sense that each household has its own controller and all the computations are performed locally. The controllers that use information only from its own household (the H variants) do not need any outside communication. The other controllers need only one-way communication from the grid to the controller. It brings the advantage that controllers can be applied without the need to change the infrastructure of the grid. Moreover, it also better preserves the privacy of the owners of the electric vehicles, as the charging requests are not shared outside of the household.

All the controllers are memory efficient and fast to compute. The NN-based variants only need to store the weight matrix (roughly a hundred numbers) and the internal state, which again contains only hundreds of numbers (mostly the past consumptions) that are used to compute the inputs of the control module itself. The ESN-based controller requires more memory, mostly for storing the matrix of recurrent weights in the reservoir, which accounts for 10,000 numbers with the settings used in the experiments. Overall, the NN-based controllers need only a few kB of memory and the ESN-based ones need less than 100kB. The computations require a few hundred arithmetical operations for the computation of the inputs of the control module and then a few vector-matrix multiplications in each time step. This makes the controllers suitable for implementation in cheap embedded hardware. 

The results show that the controllers significantly improve the load profile of the grid. While the charging increases the overall consumption, the maximum consumption remains almost the same. At the same time, the minimum consumption is increased. The differences between the controllers themselves are relatively small, but the controllers that use information about the overall consumption have slightly better results than those than do not use this information. On the other hand, the advantage of the latter controllers is that they can be used without changing anything else in the grid. We also compared two optimization algorithms and found out that controllers optimized by the CMA-ES algorithm provide better performance than those optimized by the gradient method.

We have posed the problem of controller optimization as a reinforcement learning problem. However, if we would assume (for training) that all the requests and future consumptions are known, we could obtain for each time step and each household the optimal charging current such that the objective is minimized using quadratic programming. Using the optimal charging currents, it would be possible to reformulate the problem as a supervised learning problem, where the controllers learn to charge with the optimal speed. This is left as a future work. We already made some experiments with the application of the gradient method to the best solutions given by the CMA-ES algorithm, and while there did not seem to be any significant improvements, we believe the hybridization of these approaches can bring interesting results. 

\section*{Acknowledgment}

This work was supported by Czech Science Foundation project no.~17-10090Y.

Access to computing and storage facilities owned by parties and projects
contributing to the National Grid Infrastructure MetaCentrum, provided under the
programme "Projects of Large Research, Development, and Innovations
Infrastructures" (CESNET LM2015042), is greatly appreciated.

%\IEEEtriggeratref{8}

\bibliographystyle{IEEEtran}
\bibliography{biblio.bib}

\end{document}